\documentclass[runningheads]{llncs}
\usepackage{graphicx}
\usepackage{booktabs}
\usepackage{multirow}
\usepackage{algorithm}
\usepackage{bbm}
\usepackage{wrapfig,lipsum,booktabs}
\usepackage{algorithmic}
\usepackage{xcolor}
\usepackage[T1]{fontenc}

\begin{document}
\title{\textsc{EAGLE}: A Domain Generalization Framework for AI-generated Text Detection}
\titlerunning{EAGLE: AI-generated Text Detection}
%
\author{Amrita Bhattacharjee\inst{1} \and
Raha Moraffah\inst{1} \and
Joshua Garland\inst{1} \and
Huan Liu \inst{1}}

%
\authorrunning{A. Bhattacharjee et al.}
%
\institute{Arizona State University, Tempe AZ 85281, USA \email{\{abhatt43,rmoraffa,joshua.garland,huanliu\}@asu.edu}} 
%
\maketitle              
\begin{abstract}
With the advancement in capabilities of Large Language Models (LLMs), one major step in the responsible and safe use of such LLMs is to be able to detect text generated by these models. While supervised AI-generated text detectors perform well on text generated by older LLMs, with the frequent release of new LLMs, building supervised detectors for identifying text from such new models would require new labeled training data, which is infeasible in practice. In this work, we tackle this problem and propose a domain generalization framework for the detection of AI-generated text from unseen target generators. Our proposed framework, \textsc{EAGLE}, leverages the labeled data that is available so far from older language models and learns features invariant across these generators, in order to detect text generated by an unknown target generator. 
\textsc{EAGLE} learns such domain-invariant features by combining the representational power of self-supervised contrastive learning with domain adversarial training.
Through our experiments we demonstrate how \textsc{EAGLE} effectively achieves impressive performance in detecting text generated by unseen target generators, including recent state-of-the-art ones such as GPT-4 and Claude, reaching detection scores of within 4.7\% of a fully supervised detector.

\end{abstract}

\section{Introduction}

Large language models (LLMs) are becoming ubiquitous and an increasing number of people are using these models, often equipped with easy-to-use public facing APIs, for a variety of use cases. Being brilliant productivity aids, these systems are being used for creative writing, homework help, education, and general writing assistants. However, given the human-like quality of text generated by these models, these are also susceptible to being used maliciously to generate misinformation, disinformation and misleading content at scale~\cite{tamkin2021understanding,vykopal2023disinformation,goldstein2023can}. 
Such misuse is even more harmful during major social or political events such as presidential elections~\cite{Swenson_Chan_2024,Klepper_Swenson_2023a}. Furthermore, with improvement in model fluency, recent work has shown that human readers often struggle to differentiate between actual human written text and LLM generated text~\cite{clark2021all}. This necessitates the development of automated systems to detect such LLM-generated\footnote{or `AI-generated', used interchangeably throughout this paper.} content.

Most existing work on automated detectors for AI-generated text are supervised classifiers trained using labeled data from some text generator(s), but these detectors do not generalize well to newly released, possibly much larger and more capable LLMs ~\cite{pu2023zero}. However, given the sheer variety of LLMs available alongside the fast development and release of new LLMs, it is challenging to develop, train and maintain a general purpose AI text detector that would also work for newly emerging LLMs, since this require continuous collection and/or generation of new labeled training datasets. While recently there have been some zero-shot detection methods proposed ~\cite{mitchell2023detectgpt,hans2024spotting}, most of these methods
rely on a proxy language model, and are therefore highly sensitive to the choice of this model. Furthermore, these methods do not make use of the available labeled data from older language models. In this work, we aim to leverage the labeled data that is already available, and we propose a novel domain generalization framework to learn domain-invariant (i.e., LLM-invariant) features to perform detection on text generated from a completely unseen domain (i.e., a new LLM). We assume that existing data from older generators must contain crucial discriminative features that we can leverage in a detection framework. Our proposed framework aims to capture the (1) cross-domain invariance: that is, learning the invariant features across different generators, and (2) in-domain invariance: that is, learning better latent representations in order to be robust against minor perturbations in text. For the purposes of this paper, we focus on the critical issue of AI-generated news articles and we demonstrate the effectiveness of our proposed framework on established benchmark datasets as well as data from newer LLMs, including our own GPT-4 generated data. Overall, in this work, our contributions are:

\begin{itemize}
    \item We propose \textsc{EAGLE}\footnote{\textit{inspired by the sharp eyesight of this specific bird of prey}}: a novel domain generalization for AI-generated text detection framework to detect text from new, unseen target generators, by leveraging labeled data from pre-existing, possibly older generators.
    \item Through comprehensive experiments on text from a variety of language models, we demonstrate the effectiveness of \textsc{EAGLE} in learning domain-invariant features.
    \item Alongside generating our own GPT-4 data, we evaluate the efficacy of \textsc{EAGLE} on detecting text from new state-of-the-art LLMs, by only leveraging data from older, much smaller language models.
\end{itemize}


\section{Related Work}
\label{sec:related}

\textbf{AI-generated Text Detection. }  With the rapid progress of language models over the last several years, there have also been approaches proposed for detection of text generated by such models. In the case where plenty of labeled data is available, fine-tuned pre-trained language models are often the best performing detectors ~\cite{ippolito2019automatic,he2023mgtbench}. An example of this is the OpenAI detector that is simply a RoBERTa~\cite{liu2019roberta} model fine-tuned on GPT-2 data~\cite{solaiman2019release}. A recent supervised method called Ghostbuster uses a series of weaker models followed by a search over combination functions and then a linear classifier~\cite{verma2023ghostbuster}. Some recent work also explore the use of LLMs as the detector for detecting AI-generated text~\cite{bhattacharjee2023fighting}. Authors in ~\cite{bhattacharjee2023conda} propose an unsupervised domain adaptation framework to detect AI-generated text by leveraging labeled source and unlabeled target data.

\textbf{Zero-shot AI-generated Text Detection. } In addition to supervised methods for detection, recently there has been a lot of effort in the area of zero-shot detection of text generated by AI text generators, i.e., LLMs. While some works~\cite{pu2023zero,mireshghallah2023smaller} analyze the zero-shot transfer capabilities of AI-text detectors to text from new generators, some also propose novel zero-shot or unsupervised detection methods. Some of these methods ~\cite{gehrmann2019gltr,bakhtin2019real} leverage statistical measures to identify generation artifacts across common sampling schemes, while some rely on assumptions surrounding the log probabilities of the generated text under a proxy model ~\cite{mitchell2023detectgpt,bao2023fast,su2023detectllm}. Under a full black-box setting, with no access to the token probabilities, a recent approach ~\cite{yang2023dna} uses n-gram analysis to detect such AI-generated text. Another interesting recent zero-shot detection method compares the perplexity of the input text under two related language models as a signal towards detection~\cite{hans2024spotting}. Authors in ~\cite{bhattacharjee2023conda} proposes a domain adaptation framework for unsupervised detection. Zero-shot detection approaches have also been proposed for code generated by LLMs ~\cite{yang2023zero}. 

While these various approaches have shown promising results in the detection of AI-generated text both in fully supervised and zero-shot scenarios, there is currently no work that leverages labeled data from older generators to perform unsupervised detection of text from newer generators. To the best of our knowledge, this is the first work to propose a domain generalization framework for AI-generated text detection, whereby we investigate if we can tackle the real-world scenario of detecting text from an unseen generator while leveraging labeled data from older, potentially much smaller generators.

\section{Background \& Preliminaries}

Operationalizing  machine learning frameworks for real-world use requires such methods to have the ability to perform well under domain shift. Examples of such cases in text classification are product review sentiment classifiers that are trained for a specific category of products (e.g., Books)
and need to perform well to the unseen category of Electronics. There have been substantial work done to tackle such scenarios by domain adaptation techniques. Typically such domain adaptation methods assume access to one or multiple source domains as well as \textit{unlabeled data} from the target domain. The model is then trained to optimize an objective function that retains performance on the source domain while learning the invariance across the source and target domains. 

In our task of AI-generated text detection, we treat each generator or LLM as a `domain'; that is, texts generated by different LLMs follow a different distribution. In this work, we consider the case where we have access to labeled data from $k$ such generators, and we aim to detect text generator by an unseen $(k+1)$th generator. This is a typical domain generalization setting. While numerous approaches have been proposed for this setting in the field of computer vision, text is relatively under explored~\cite{jia2022prompt,krishnan2020diversity,tan2022domain}. A widely used method is DANN - Domain Adversarial Neural Networks~\cite{ganin2015unsupervised,ganin2016domain} - that uses domain adversarial training to make the model learn domain invariant features. Such a method has also been adapted to text applications for sentiment classification, albeit with unlabeled data from the target. In this work, we build on top of this framework by: (1) modifying it to only use source domain data from multiple sources and \textit{no} target data, and (2) adding an in-domain contrastive regularization function via which the model aims to learn more robust, better representations of the text. As supported by prior work~\cite{xu2012robustness}, this additional regularization component should provide more generalizability to the trained model.  

\section{Problem Definition}

In this work, we address the task of AI-generated text detection. Specifically, we assume we have access to labeled data from $k$ different AI-text generators, or LLMs, denoted by $\{D^1, D^2,\dots,D^k\}$. Each domain is defined as the combination of the input (or feature) space and the label space: $D^i = \{X^i, Y^i\}, \forall i \in [1,k]$. Typically, for most domain generalization use-cases, including our problem setting, the label space is constant across all the different domains, that is, $Y^1 = Y^2 = ... = Y^k$. The goal of domain generalization in a classification setting is to learn a mapping $f: X \rightarrow Y$ that captures the domain-invariant features across the $k$ domains in order to minimize the predictive loss on data from a new unseen domain $D^{k+1}$.

In the task of AI-generated text detection, with text from different generators, we formulate the problem as follows: 
Each generator $i$ is a domain $D^i$, $X^i$ consists of text samples either generated by the generator $i$ or written by a human, the label space $Y = \{0,1\}$ denoting human-written and AI-generated text respectively. Hence the function $f(\cdot)$ that we aim to learn should ideally capture the features invariant across different text generators, while being discriminative enough to be able to distinguish between generated vs. human-written text. In the following section, we describe our proposed framework to learn this function to perform the task of AI-generated text detection on unknown target domains.

\section{Proposed Framework}
\label{sec:framework}

\begin{figure}
    \centering
    \includegraphics[width=\columnwidth]{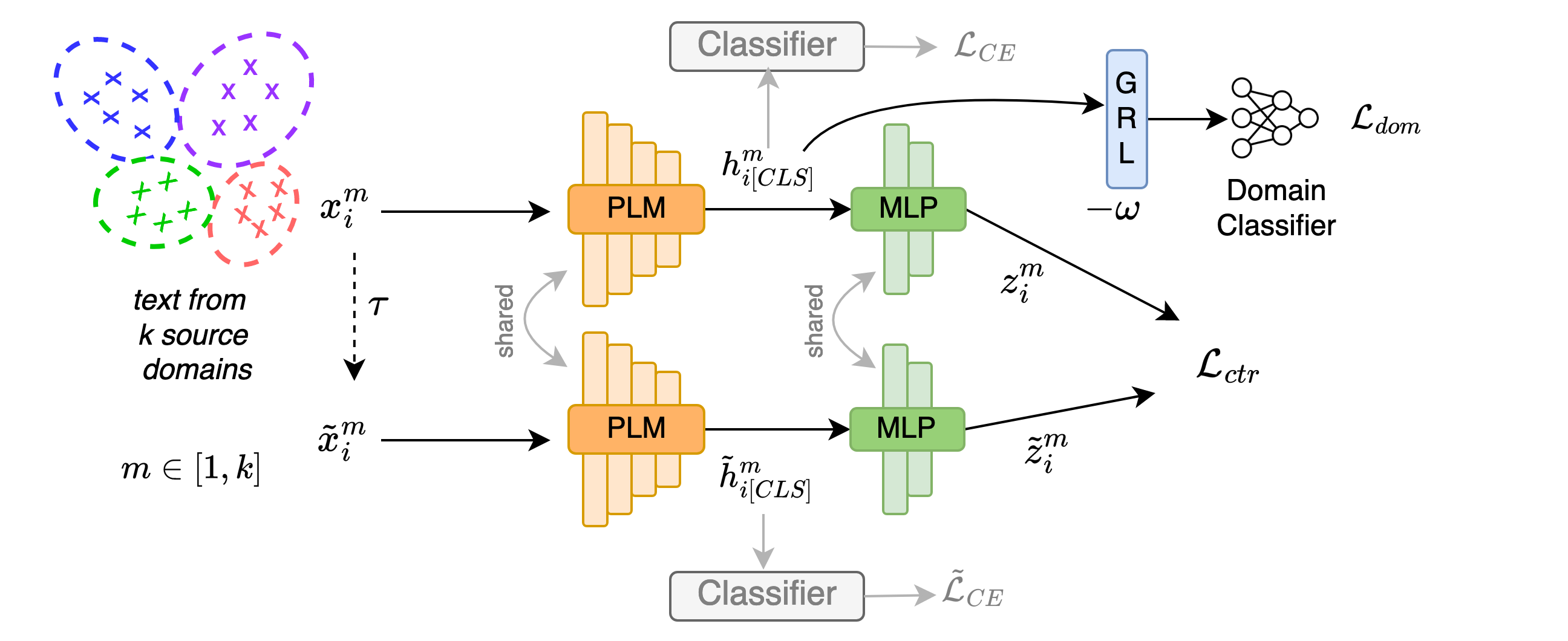}
    \caption{Our proposed EAGLE framework.}
    \label{fig:framework}
\end{figure}

In this section, we introduce our proposed \textsc{EAGLE} framework, as showed in Figure \ref{fig:framework}, and describe each component in detail. 

\textbf{Classification Backbone. } Given the superior performance of pre-trained language models on a variety of tasks, and the ease of fine-tuning such models on task-specific data, we opt to use a pre-trained language model (PLM) RoBERTa~\cite{liu2019roberta} (roberta-base) from Huggingface transformers\footnote{https://huggingface.co/FacebookAI/roberta-base}, along with a classifier head on top of it. Our input text consists of labeled text data from $k$ different domains: $x^m_i \in \{D^m\}_{m=1}^{k}$. Each text $x^m_i$ is fed into the PLM and we obtain the final hidden layer embedding $h^{m}_{i[CLS]}$. Since we have the labeled data from each of the $k$ sources, we feed this embedding into the linear classifier layer and compute the binary-cross entropy loss $\mathcal{L}_{CE}$. Since we also have a perturbed version of the input (as explained later in the contrastive loss section), we compute a similar loss for this and denote it as $\tilde{\mathcal{L}}_{CE}$. 

\begin{equation}
    \mathcal{L}_{CE}  = - \frac{1}{b} \sum_{i=1}^b [y_i \log p(y_i|h^m_{i[CLS]}) + (1-y_i) \log(1-p(y_i|h^m_{i[CLS]}))]
\end{equation}

\noindent where $y_i$ is the ground truth label for input $x_i^m$ and $b$ is the batch size. We have a similar loss for the perturbed version of the text as well.

\textbf{Domain Adversarial Training. } For the task of domain generalization, we would require our framework to learn features that are invariant across the different domains, i.e., text generators. To facilitate this, we employ domain adversarial training~\cite{ganin2015unsupervised,ganin2016domain}. To do this, we use the final hidden layer embedding $h^{m}_{i[CLS]}$ for each input text, and feed this into a domain classifier that computes a domain loss. The objective of the domain classifier is to minimize the classification error of distinguishing the source and target domains, while the objective of the entire classification model is to learn domain-invariant i.e., transferable features. Through this adversarial training process, the model learns better, more discriminative representations of the input text specific to the downstream task that are also domain-invariant. To facilitate gradient-based optimization of these conflicting objectives, a gradient reversal layer (GRL) is used~\cite{ganin2016domain}. More precisely, the hidden layer embedding $h^{m}_{i[CLS]}$ is fed into the gradient reversal layer, and then into the domain classifier. The gradient reversal layer does not have any trainable parameters. In the forward pass, it essentially functions as an identity transform, and in the backpropagation, the GRL layer reverses the gradient by multiplying the gradient from the subsequent layer by a negative scalar value $-\omega$. Therefore, in the forward pass, the output of this layer is

\begin{equation}
    GRL(h^{m}_{i[CLS]}) = h^{m}_{i[CLS]}
\end{equation}

This is now input into the domain classifier which is simply a dropout layer followed by a linear layer with dimension $768 \times num\_domains$, where $num\_domains$ is the number of source domains we are using. Say the predicted domain label for this input is $d_i^m$. We compute the domain loss as the cross-entropy loss as

\begin{equation}
    \mathcal{L}_{dom} = - \frac{1}{k} \sum_{m=1}^{k} \sum_{i=1}^{b} d_i^m \log p(d_i^m)
\end{equation}

\noindent where $k$ is the number of source domains, $b$ is the batch size. During backpropagation of this loss $L_{dom}$ through the GRL layer, the gradient becomes:

\begin{equation}
    GRL(\frac{\partial L_{dom}}{\partial \theta_f}) = - \omega \frac{\partial L_{dom}}{\partial \theta_f}
\end{equation}

\noindent where $\theta_f$ are the parameters of the feature extractor network, which is the pre-trained RoBERTa language model in our case, and $- \omega$ is the negative scalar with which the gradient is reversed. Training the model using the GRL ensures that the domain classifier is optimized to predict domain labels correctly, while the RoBERTa parameters are optimized to deceive the domain classifier, thereby learning domain invariant features.

\textbf{Contrastive Learning for Better Representations. }  Inspired by previous work, we add a contrastive loss component to learn better representations of the input text. A contrastive loss component would act as a regularizer to learn more robust representations of the input. Ideally our framework should be robust to small perturbations in the input, to facilitate in-domain invariance to such noisy perturbations. To facilitate that, we use a loss such as in ~\cite{bhattacharjee2023conda}. Following previous work, for each input, we apply a perturbation $\tau$ which is synonym replacement. We use the hidden layer embeddings of the original and perturbed texts: $h^{m}_{i[CLS]}$ and $\tilde{h}^{m}_{i[CLS]}$, and pass these through a projection layer in order to compute the contrastive loss in a lower dimensional space, as done in ~\cite{bhattacharjee2023conda}, and compute a SimCLR-style~\cite{chen2020simple} contrastive loss between the original and the perturbed views of the input text. Here the original and perturbed views are the positives, and all other samples in the mini-batch are considered as negatives. Therefore, the contrastive loss is

\begin{equation}
{\mathcal{L}_{ctr} = - \sum_{i \in b} \log \frac{exp(sim(z^m_i,\tilde{z}^m_i)/t)}{\sum_{j=1}^{2|b|} \mathbbm{1}_{[j \neq i]} exp(sim(z^m_i,z_j)/t)}}
\end{equation}

\noindent where $z^m_i$ and $\tilde{z}^m_i$ refer to the projected embeddings in the lower dimension space, corresponding to the original and the perturbed views of the input text respectively, $t$ is the temperature. 

The final training objective is

\begin{equation}
    \mathcal{L} = \frac{\lambda_1}{2} (\mathcal{L}_{CE} + \tilde{\mathcal{L}}_{CE}) + \lambda_2 \mathcal{L}_{ctr} + \lambda_3 \mathcal{L}_{dom}
\end{equation}

\noindent where $\lambda_1, \lambda_2$ and $\lambda_3$ are hyper-parameters, $\tilde{\mathcal{L}}_{CE}$ is the cross-entropy loss from the perturbed version of the text.

\section{Experimental Settings}
In this section, we describe the datasets we trained and evaluated our framework on, along with the baselines and experimental settings used. 

\subsection{Datasets}
\label{sec:data}

In this work, we specifically focus on AI-generated news articles. The nature of our domain generalization task also requires having AI-generated text from separate AI-text generators (i.e., language models or `domains'). Therefore, in our experiments, we use the benchmark dataset TuringBench. TuringBench~\cite{uchendu2021turingbench} is an extensive dataset consisting of new-style text from a total of 19 different generators along with human-written text. These 19 generators comprise various sizes of 10 different model architectures: \{GPT-1~\cite{radford2018improving}, GPT-2~\cite{radford2019language}, GPT-3~\cite{brown2020language}, GROVER~\cite{zellers2019defending}, CTRL~\cite{keskar2019ctrl}, XLNET~\cite{yang2019xlnet}, XLM~\cite{lample2019cross},  TRANSFORMER-XL~\cite{dai2019transformer}, FAIR~\cite{ng2019facebook,chen2020facebook}, and PPLM~\cite{dathathri2019plug}\}.
Following previous work~\cite{bhattacharjee2023conda}, we use a subset of 6 of these generators as our domains. This choice ensures that we have a good coverage of different architectures and model families of generators. The generators are:

\noindent
\textbf{CTRL}: This is a 1.5B parameter transformer-based language model that is capable of controlled generation of text. Generation can be conditioned on control codes such as style, sentiment, etc. Authors in ~\cite{uchendu2021turingbench} generate text from CTRL using the \textit{News} control code.

\noindent
\textbf{FAIR\_wmt19}: This is a 656M parameter transformer-based model developed by FAIR as part of their submission to the WMT19 news translation task. Texts are generated using the English version of the model using the FAIRSEQ~\cite{ott2019fairseq} toolkit.

\noindent
\textbf{GPT2\_xl}: This is the 1.5B version of the GPT-2 model, the precursor to GPT-3 and the more recent GPT-3.5 and GPT-4 models. 

\noindent
\textbf{GPT-3}: This is the largest model in the TuringBench dataset. It is a 175B model which is a successor of the previous GPT-2 model.

\noindent
\textbf{GROVER\_mega}: This is a 1.5B parameter transformer-based model, which is trained to generate news style text. 

\noindent
\textbf{XLM}: This 550M parameter model is the smallest model we use in our evaluation. This is a transformer-based model that is designed for cross-lingual tasks.

Given the rapid development and release of new text generators or LLMs, it is necessary to evaluate AI-text detection systems on new language models, that are not covered by the TuringBench dataset. Therefore, we also use news-style data generated by new language models such as \textbf{GPT-3.5}, \textbf{GPT-4} from OpenAI~\cite{achiam2023gpt} and \textbf{Claude} from Anthropic~\cite{bai2022constitutional}. GPT-3.5 (or often referred to as ChatGPT) and its newer variant GPT-4 have garnered immense attention from academics, industry, educators and practitioners for its impressive instruction-following and text generation capabilities, as well as superior performance on complex tasks. Given the human-like quality of text generated by GPT-3.5 and GPT-4, there have also been concerns surrounding whether such text can be detected by automated AI-text detection tools. To investigate this, we use such data in our evaluation. For GPT-3.5, we use the data from ~\cite{bhattacharjee2023conda}. For Claude, we use data from ~\cite{verma2023ghostbuster}. For GPT-4, we generate our own news-style data as described below. 

Following a similar data generation pipeline as in ~\cite{bhattacharjee2023conda}, we first collect a set of $2,000$ human written articles from CNN and Washington Post (as done in previous work ~\cite{uchendu2021turingbench}). For each article, we use the \texttt{headline} of the article to generate an article using GPT-4 by simply prompting the model with the following prompt: ``Generate a news article with the headline `<\texttt{headline}>'."
Due to resource constraints, we only generate $2,000$ articles using GPT-4. Therefore, for GPT-4, we have a balanced set of $2,000$ human-written articles and $2,000$ GPT-4 generated articles.

\subsection{Baseline Detection Methods}
\label{sec:baselines}

Similar to previous work~\cite{bhattacharjee2023conda,hu2023radar}, we use a variety of unsupervised baselines: 

\textbf{GLTR}~\cite{gehrmann2019gltr}: This is a set of four statistical measures to identify whether a text is AI generated or not. These are computed based on token-wise log probabilities. These measures are: (1) log probability ($\log p(x)$): that assumes that higher log probability indicates that the text is AI-generated, (2) average token rank and (3) token log-rank: that are based on the assumption that tokens with lower rank are possibly AI-generated, and (4) predictive entropy: that is based on the assumption that AI-generated text often has less diversity than human-written, and therefore less entropy.  

\textbf{DetectGPT}~\cite{mitchell2023detectgpt}: This method relies on a proxy language model to compute log probabilities of generated tokens. The main assumption in this method is the minor perturbations of AI-generated text would fall in the negative curvature region of the log-likelihood curve, while such perturbations to human-written text does not follow this trend. This assumption on the curvature of the log probabilities is used as the discriminative feature for prediction.

Apart from these unsupervised baselines, in order to test the transferability across different domains (i.e., generators) we also use standard baselines as in ~\cite{pu2023zero}: 

\textbf{Data Mix}: Here we combine the data from the training splits of the $k$ known generators, and train one detector on this unified dataset. During evaluation, we evaluate this trained model on the test split of the unseen $(k+1)$th generator. 

\textbf{Ensemble}: Here for each of the $k$ known generators, we train a separate detector, and then during inference on the unknown $(k+1)$th generator, we evaluate all $k$ detectors, and use an aggregation of the $k$ prediction scores to get the final label. For the aggregation, we use max and average. 

For both Data Mix and Ensemble baselines, we use a RoBERTa model (roberta-base) with a classification head on top as the detector. We train this for 3 epochs with Adam optimizer~\cite{kingma2014adam} and a learning rate of $2\times10^{-5}$.



\subsection{Implementation Details}

All our experiments were performed using PyTorch, on an NVIDIA A100 GPU with 40 GB memory. For hyperparameter values in the detection experiments, we use grid search and use the best performing value in our final experiments. For generating the GPT-4 data, we use temperature of 0.5 and top\_p of 1. To facilitate reproducibility, all code and data will be made available at <link-to-be-inserted-after-blind-review>.

\begin{table*}[]
\centering
\resizebox{\textwidth}{!}{%
\begin{tabular}{@{}ccccccccccc@{}}
\toprule
\multirow{2}{*}{\begin{tabular}[c]{@{}c@{}}\textbf{Target} \\ \textbf{Generator}\end{tabular}} &
  \multirow{2}{*}{\begin{tabular}[c]{@{}c@{}}\textbf{Fully} \\ \textbf{Supervised} \end{tabular}} &
  \multicolumn{4}{c}{\textbf{GLTR}} &
  \multirow{2}{*}{\textbf{Det.GPT}} &
  \multirow{2}{*}{\textbf{DataMix}} &
  \multicolumn{2}{c}{\textbf{Ensemble}} &
  \multirow{2}{*}{\textbf{Ours (k=5)}} \\ \cmidrule(lr){3-6} \cmidrule(lr){9-10}
             &       & \textbf{log p(x)} & \textbf{rank}  & \textbf{log rank} & \textbf{entropy} &        &       & \textbf{Avg}   & \textbf{Max}   &       \\ \midrule
CTRL         & 1     & 0.951    & 0.849 & \underline{0.956}    & 0.379   & 0.793  & 0.902 & 0.801 & 0.761 & \textbf{0.984} \\
FAIR\_wmt19  & 0.999 & 0.558    & 0.618 & 0.546    & 0.656   & 0.5045 & 0.797 & \underline{0.836} & 0.728 & \textbf{0.896} \\
GPT2\_xl     & 0.998 & 0.485    & 0.508 & 0.48     & 0.631   & 0.529  & \textbf{0.995} & \underline{0.993} & 0.982 & 0.94  \\
GPT3         & 0.988 & 0.362    & 0.356 & 0.341    & 0.756   & 0.5485 & \underline{0.976} & 0.951 & 0.947 & \textbf{0.983} \\
GROVER\_mega & 0.996 & 0.434    & 0.469 & 0.434    & 0.592   & 0.5415 & \underline{0.895} & 0.702 & 0.705 & \textbf{0.912} \\
XLM          & 1     & 0.473    & 0.762 & 0.442    & 0.696   & 0.7355 & 0.864 & \underline{0.978} & 0.952 & \textbf{0.98}  \\ \bottomrule
\end{tabular}%
}
\vspace{0.4cm}
\caption{Performance of our framework on TuringBench dataset. Scores are AUROC values averaged over three different seeds. Best values are in \textbf{bold} and second best values are \underline{underlined}. Det.GPT refers to the DetectGPT baseline~\cite{mitchell2023detectgpt}.}
\label{tab:main}
\end{table*}

\section{Experimental Results}

To investigate the effectiveness of our \textsc{EAGLE} framework, we perform a comprehensive set of experiments to answer the following research questions:

\begin{itemize}
    \item \textbf{RQ1:} Can our framework learn domain-invariant features and perform well on unseen generators?
    \item \textbf{RQ2:} Can our framework learn to detect text from new generators after being trained only on text from older generators?
\end{itemize}

Apart from these main research questions, we also explore the effectiveness of the different components in the framework, as well as the effect of number of source domains.

\subsection{RQ1: Performance on Unseen Targets}



We evaluate our framework on the TuringBench dataset and report these results in Table~\ref{tab:main}. Here we compare the performance of our framework with unsupervised baselines as described in Section \ref{sec:baselines}. We also compare performance with a fully supervised detection model that serves as an upper bound. For the fully supervised model we use a RoBERTa~\cite{liu2019roberta} (roberta-base\footnote{https://huggingface.co/FacebookAI/roberta-base}) model with a classification head on top and the model is fine-tuned on labeled data from the target generator. This essentially serves as an upper bound for the performance. We 
report the AUROC scores from this experiment in Table \ref{tab:main}. We see that for all of the generators in our experiment, except GPT2\_xl, our proposed framework performs better than the unsupervised baselines, with no labeled target data.

\begin{figure}
    \centering
    \includegraphics[width=0.7\textwidth]{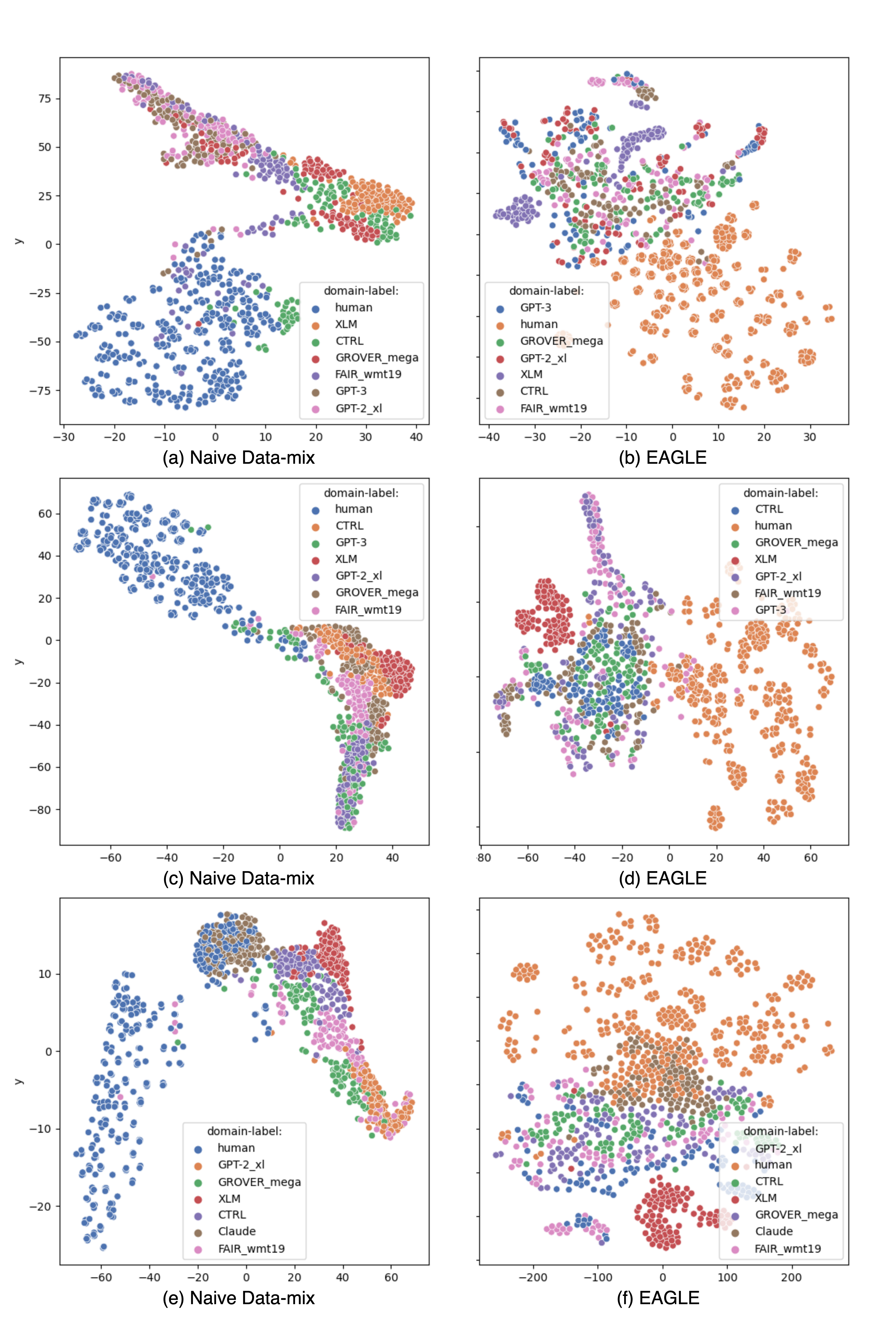}
    \caption{t-SNE visualizations for representations learned by our EAGLE model, vs the naive data-mix baseline with the same set of sources. (a)-(b) denote plots for data mix and EAGLE representations for target generator CTRL, respectively, (c)-(d) for target generator GPT-3 and (e)-(f) for target generator Claude.}
    \label{fig:tsne}
\end{figure}

To understand if our framework is effectively learning the domain invariant features, we visualize the t-SNE plots~\cite{van2008visualizing} of the learned representations from our EAGLE framework vs. a naive data mix framework with the same set of source domains
, for a fair comparison. We produce these visualizations for three target generators (top row to bottom row): CTRL, GPT3 and Claude, and show these in Figure \ref{fig:tsne}. For each target generator, we see that the data mix plot (left) shows somewhat homogeneous clusters for each of the generators, implying texts from these different source generators and the target generators lie in disjoint spaces in the latent representation space. However, for the EAGLE plots (right), we see more overlap between text from the different generators. This is encouraging since this implies our framework has successfully learned domain-invariant features, thereby making the data points inseparable based on the domain label. For Claude, we see some overlap between human written articles and Claude generated ones, implying that Claude-generated text is still somewhat challenging to distinguish from human-written text.


\begin{table}[]
\centering
{
\begin{tabular}{@{}ccccccc@{}}
\toprule
\multirow{2}{*}{\textbf{Target}} & \multicolumn{2}{c}{\textbf{Data Mix}} & \multicolumn{2}{c}{\textbf{Ensemble}} & \multicolumn{2}{c}{\textbf{EAGLE (Ours)}} \\ \cmidrule(l){2-7} 
        & Avg            & Max & Avg            & Max & Avg            & Max \\ \midrule
Claude  & 36.67 $\pm$ 22.18 & 69  & 0.4 $\pm$ 0.43    & 0.8 & \textbf{40.67 $\pm$ 29.85} & \textbf{97}  \\
GPT-3.5 & 75 $\pm$ 19.79   & 98  & 54.72 $\pm$ 45.22 & 96  & \textbf{86.17 $\pm$ 14.66} & \textbf{99}  \\
GPT-4   & 32.5 $\pm$ 20.98  & 64  & 1.62 $\pm$ 1.63 & 3.1 & \textbf{58 $\pm$ 37.18}   & \textbf{99}  \\ \bottomrule
\end{tabular}
}
\vspace{0.4cm}
\caption{Performance of our model on data from newer generators. Values are TPR. We report both the average across all possible sets of 5 sources (average $\pm$ standard deviation) and the maximum across the choices. Best performance is in \textbf{bold}.}
\label{tab:new-llm}
\end{table}


\begin{table}
\centering
\begin{tabular}{@{}cccc@{}}
\toprule
\begin{tabular}[c]{@{}c@{}}\textbf{Framework} \\ \textbf{variant}\end{tabular} & \textbf{CTRL}  & \textbf{GRO\_m} & \textbf{XLM}   \\ \midrule
\textsc{EAGLE}                                                         & \textbf{0.984} & \textbf{0.912}        & \textbf{0.980}  \\
\textsc{EAGLE}  $ - \mathcal{L}_{ctr}$                                            & 0.763 & 0.883        & 0.932 \\
\textsc{EAGLE} $ - \mathcal{L}_{dom}$                                               & 0.948 & 0.814        & 0.968 \\
\textsc{EAGLE}  $ - \mathcal{L}_{ctr} - \mathcal{L}_{dom}$                                                 & 0.902 & 0.895        & 0.970 \\ \bottomrule
\end{tabular}
\vspace{0.4cm}
\caption{Performance comparison across different variants of \textsc{EAGLE}. GRO\_m refers to GROVER\_mega. Scores are AUROC and best performance is in \textbf{bold}.}
\label{tab:abla}
\end{table}

\subsection{RQ2: Detection of Text Generated by Newer LLMs}



Here we are interested in evaluating the performance of our framework on new state-of-the-art LLMs: Claude, GPT-3.5 and GPT-4, \textit{without} using data generated by any of these in the training set. For this experiment, we use the same framework and $k=5$ number of source domains. For each new LLM (Claude, GPT-3.5 and GPT-4), we only consider $k=5$ \textit{older} generators as the source domains. This source domain data is the same TuringBench data as used in the previous experiment. Essentially, we are interested in evaluating how well our proposed framework can transfer discriminative features from older generators with possibly easily available data, to newer generators from which text is more difficult to identify. We report the true positive rate of detection in Table \ref{tab:new-llm} from the best performing set of sources, along with the average across different 5-set choices. We also compare results from our framework with two baseline methods: (1) Data Mix - where the training data comes from $k=5$ source domains (from generators in TuringBench), and (2) Ensemble - where $k=5$ models, each trained on a single source are used. We keep the same set of sources across all 3 settings: our framework, data mix and ensemble. This is because we are interested in understanding how our framework can learn domain invariant features from the same data, above and beyond what is already do-able by using all of the same sources in a naive way.

\begin{wrapfigure}{R}{0.5\textwidth}
    \begin{center}
    \includegraphics[width=0.5\textwidth]{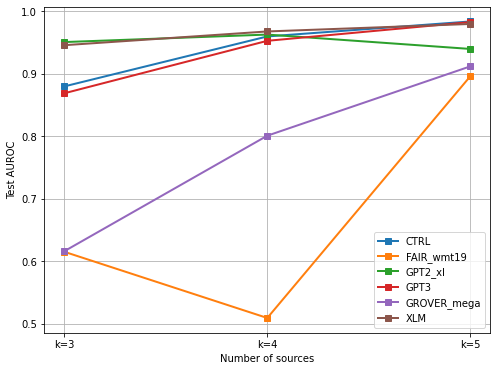}
    \end{center}
    \caption{Variation of test performance with respect to number $k$ of source domains or generators used in training.}
    \label{fig:k-test}
\end{wrapfigure}

We report the true positive rate of detection of LLM-generated text for both `avg' and `max' settings: where we take the average and maximum across all combinations of 5 out of the 6 sources, respectively (as described in Section \ref{sec:data}). We see that our framework performs better than both the Data Mix and Ensemble baselines by a significant margin. Interesting, we see that the Ensemble approach performs the worst, and has negligible performance for detecting Claude and GPT-4 generated text. However, performance varies significantly across different choices of sources as depicted by the high values of standard deviation in most cases. 

\subsection{Ablation: Effectiveness of Framework Components}

In this experiment, we evaluate the effectiveness of the different components in our framework. We remove one component at a time, train and evaluate each variant and report these results for a randomly chosen set of three target domains in Table \ref{tab:abla}. \textsc{EAGLE}  $ - \mathcal{L}_{ctr}$ removes the contrastive loss component (along with the cross-entropy loss for the perturbed version of the input text), \textsc{EAGLE}  $ - \mathcal{L}_{dom}$ removes the domain loss, and \textsc{EAGLE}  $ - \mathcal{L}_{ctr} - \mathcal{L}_{dom}$ removes both. This last variant is essentially similar to the Data Mix baseline. We see performance drops for each of those variants, while $\mathcal{L}_{ctr}$ and $\mathcal{L}_{dom}$ having different degrees of impact on the performance for different target generators. Overall, we see our full framework \textsc{EAGLE} performs the best across all the target generators.

\subsection{Hyperparameter Analysis: Effect of Number of Sources}

We are also interested in evaluating how sensitive our model is to the number of source generators we include in our training. For this analysis, we vary the number of source domains/generators $k = \{3,4,5\}$. We show the variation of performance across the different choices of $k$ in Figure \ref{fig:k-test}. We see that in general, the performance increases with increase in the value of $k$. This is possibly due to the increase in variability and diversity of the data as we increase $k$, since the sources are all different model families with different architectural backbones, with the exception of GPT2\_xl and GPT3, albeit differing in the type of attention used in the 2 models~\cite{brown2020language}. Therefore, we use $k=5$ as the number of source domains in all our experiments. 

\section{Conclusion and Future Work}

In this work, we propose a novel framework called \textsc{EAGLE}, to perform AI-generated text detection from unseen target generators, by effectively leveraging data from older, possibly much smaller generators. Our framework learns domain invariant features via a gradient reversal layer and adversarial training, thereby retaining task-specific discriminative features, while learning to ignore domain specific features. We demonstrate the effectiveness of the proposed framework via experiments on a vast variety of language models, ranging from smaller language models such as XLM, to recent state of the art models such as GPT-4 and Claude. We also generate our own GPT-4 data that we will make available for research purposes upon request. Our experiments show that \textsc{EAGLE} can effectively leverage data from older generators, learn transferable features and perform detection on unseen target LLMs, in an unsupervised manner. Our framework and findings pave the way for building more detectors for newly emerging LLMs, simply by leveraging data from older, smaller generators. 

Here, we assume the test data only comes from one generator. Future work could explore the more challenging setting of multiple unseen test generators, perhaps assuming some mixture of distributions over the texts. Furthermore, we only use new-style text since this is the only type of text that is widely available from a variety of different generators. Future work could also explore how our detector performs on other types of text. Another direction could be generalization in two dimensions: first over the text generators, and second, over the type of text, such as news, scientific article, finance, etc.

\section{Ethical Statement}

With the prevalence of LLMs everywhere, we are increasingly exposed to AI-generated text. While automatic detection of such AI-generated text is a first step to evaluate content authenticity, care must be taken while using such automatic systems in practice. Most of these detectors are black-box models, do not provide explanations for their predictions, and are therefore challenging to be used in practice. For high-stakes applications where users may be penalized heavily for using AI-generated content, false positives can be highly undesirable. Similarly, for applications such as evaluating content authenticity on a creative writing website, false negatives would be highly undesirable, since real human authors may potentially miss out on the credit they deserve. Furthermore, alongside flagging generated content, systems also need to identify the intent behind the content generation - whether it is being used maliciously or not. However, such determination is highly subjective and therefore a vastly unsolved issue. For the purposes of this work, we do not condone misuse of our proposed detection framework especially for high-stakes applications and urge users of such applications to instead resort to a human-in-the-loop, hybrid solution.

\section*{Acknowledgements}

This work is supported by the DARPA SemaFor project (HR001120C0123), Army Research Lab (W911NF2020124) and Army Research Office (W911NF211 0030). The views, opinions and/or findings expressed are those of the authors.
%
%
%
%
\bibliographystyle{splncs04}
\bibliography{references}

\end{document}